\documentclass[10pt,twocolumn,letterpaper]{article}

\usepackage[accsupp]{axessibility}  
\usepackage{wacv}              

\usepackage{graphicx}
\usepackage{amsmath}
\usepackage{amssymb}
\usepackage{booktabs}
\usepackage{multirow}
\usepackage[pagebackref=true,colorlinks,bookmarks=false,citecolor=blue,linkcolor=blue,urlcolor=blue]{hyperref}
\usepackage{color}

\usepackage[capitalize]{cleveref}
\crefname{section}{Sec.}{Secs.}
\Crefname{section}{Section}{Sections}
\Crefname{table}{Table}{Tables}
\crefname{table}{Tab.}{Tabs.}

\def\wacvPaperID{1337} 
\def\confName{WACV}
\def\confYear{2024}

\begin{document}

\title{Open-NeRF: Towards Open Vocabulary NeRF Decomposition}

\author{Hao Zhang,  Fang Li,  Narendra Ahuja\\
University of Illinois Urbana-Champaign\\
{\tt\small \{haoz19, fangli3, n-ahuja\}@illinois.edu}
}
\maketitle

\begin{abstract}
   In this paper, we address the challenge of decomposing Neural Radiance Fields (NeRF) into objects from an open vocabulary, a critical task for object manipulation in 3D reconstruction and view synthesis. Current techniques for NeRF decomposition involve a trade-off between the flexibility of processing open-vocabulary queries and the accuracy of 3D segmentation. We present, Open-vocabulary Embedded Neural Radiance Fields (Open-NeRF), that leverage large-scale, off-the-shelf, segmentation models like the Segment Anything Model (SAM) and introduce an integrate-and-distill paradigm with hierarchical embeddings to achieve both the flexibility of open-vocabulary querying and 3D segmentation accuracy. Open-NeRF first utilizes large-scale foundation models to generate hierarchical 2D mask proposals from varying viewpoints. These proposals are then aligned via tracking approaches and integrated within the 3D space and subsequently distilled into the 3D field. This process ensures consistent recognition and granularity of objects from different viewpoints, even in challenging scenarios involving occlusion and indistinct features. Our experimental results show that the proposed Open-NeRF outperforms state-of-the-art methods such as LERF~\cite{lerf} and FFD~\cite{ffd} in open-vocabulary scenarios. Open-NeRF offers a promising solution to NeRF decomposition, guided by open-vocabulary queries, enabling novel applications in robotics and vision-language interaction in open-world 3D scenes. 
\end{abstract}


\begin{figure*}
  \centering
  \includegraphics[width=1\linewidth]{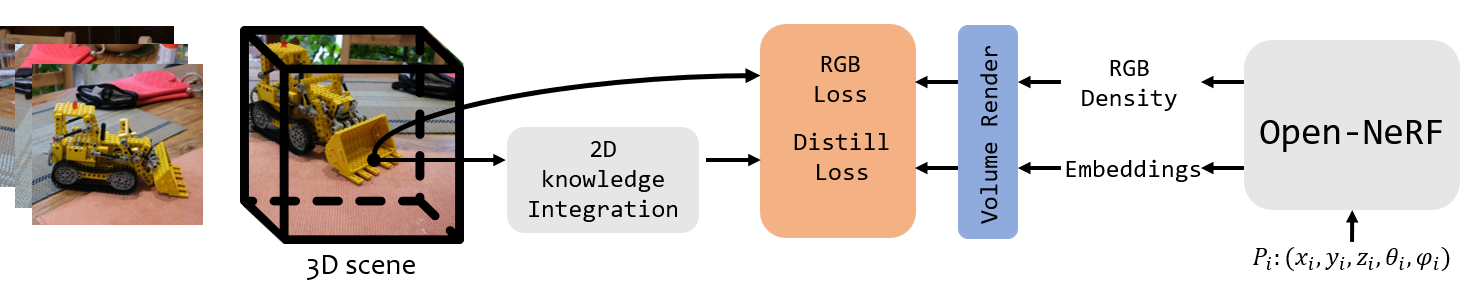}
  \caption{Overview of our Open-NeRF. We follow the integrate-and-distill paradigm, by first integrating the 2D knowledge and then distilling it into the 3D fields.}
  \label{fig3}
\end{figure*}

\section{Introduction}

\begin{figure*}
  \centering
  \includegraphics[width=1\linewidth]{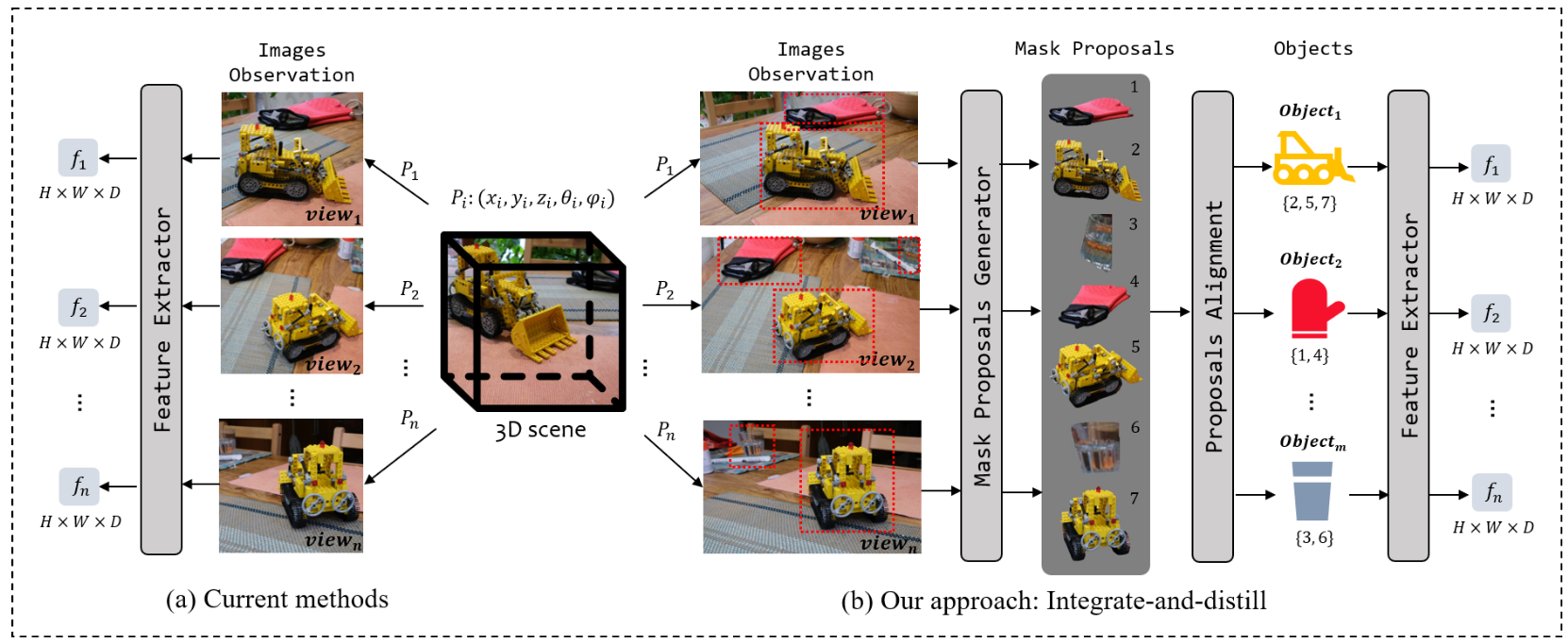}
  \caption{Comparison of existing methods and our proposed approach. (a) presents the methodology employed by existing methods like FFD~\cite{ffd} and N3F~\cite{n3f}, which involve direct extraction of per-pixel image features using off-the-shelf feature extractors such as LSeg~\cite{lseg}. These methods then distill the extracted features into 3D fields. However, these approaches often yield subpar results in certain viewpoints due to occlusions or the absence of discriminative features. (b) showcases our approach, which follows an integrate-and-distill paradigm. Initially, we generate region proposals (bounding boxes) for all objects in the 3D scene. These proposals are then input into the Segment Anything Model (SAM) to generate corresponding mask proposals and align them for the same object. Finally, we fuse all the mask proposals pertaining to the same object and extract integrated embeddings.}
  \label{fig1}
  \vspace{-5pt}
\end{figure*}

Neural Radiance Fields (NeRFs) show great promise for high-quality 3D reconstruction and novel view synthesis from 2D image observations taken from various viewpoints (camera positions and viewing angles). However, to manipulate an object in the field, such as extracting, removing, or altering its color or texture, it is essential to first decompose the 3D field.
Some methods~\cite{insnerf, dm, inplace} introduce an extra field component to learn semantic information or unique codes for all individual objects in 3D space from 2D supervision. However, these methods require annotations of ground-truth 2D masks for supervision, thus introducing additional a cost for labeling.
Other methods, such as FFD~\cite{ffd} and N3F~\cite{n3f}, address the issue using pre-trained image feature extractors like openclip-LSeg and DINO to distill 2D image features into the 3D field. However, as shown in Figure \ref{fig1} (a), they are limited to closed-set scenarios, containing objects from predefined classes, e.g., in the COCO~\cite{coco} dataset.
Some methods, like LERF~\cite{lerf}, ground language embeddings from off-the-shelf models like openclip into NeRF to locate open-vocabulary queries within 3D scenes. However, these methods lack accurate 3D segmentation, limiting their practical application.
Fortunately, with the emergence of large-scale off-the-shelf segmentation models such as the Segment Anything Model (SAM)~\cite{sam}, obtaining 2D mask proposals in open-world images has become possible. Intuitively, grounding SAM into the NeRF may help achieve both flexible open-vocabulary query processing and decent 3D segmentation accuracy. However, directly distilling the 2D knowledge obtained from the mask proposals generated by SAM into the 3D field results in undesirable outcomes due to a lack of consistency across different viewpoints.
%
As shown in Figure \ref{fig2}, (a) the recognition confidence varies with camera parameters, because some objects may be difficult to identify from certain viewing angles due to a lack of distinctive features or occlusion; (b) the granularity of masks proposals of the same object for different camera parameters is not consistent, because SAM may generate a varying number of masks proposals for the same objects in different views.

In this paper, we present a novel approach, named Open-vocabulary Embedded Neural Radiance Fields (Open-NeRFs) that introduces an integrate-and-distill paradigm. It leverages hierarchical embeddings to address issues arising from direct distilling of 2D knowledge from SAM, for flexible real-time handling of various types of open-vocabulary queries while also providing high-quality 3D segmentation results.
Instead of distilling the 2D semantic information or image embeddings from off-the-shelf models for different viewpoints
to the 3D field independently, we use the following procedure (Figure \ref{fig1}): 
(1) Leverage off-the-shelf models such as SAM to generate hierarchical 2D mask proposals for images taken from multiple viewpoints; (2) Align every mask in the 2D images with its corresponding object in the 3D space; (3) Integrate the knowledge of all 2D mask proposals for the same object across all viewpoints; (4) Distill the integrated hierarchical knowledge to the 3D field. Through this integrate-and-distill approach, we enable recognition and granularity consistency of objects from different perspectives, even when they are partially occluded or are otherwise difficult to recognize. 
To more robustly assess the performance of Open-NeRF in open-vocabulary scenarios, we incorporate an additional set of ten real-world scenes for evaluation. Experimental results demonstrate the effectiveness of our proposed approach, surpassing state-of-the-art performance in challenging open-vocabulary scenarios.
%

\section{Related Work}

\subsection{Implicit Neural Representation}
In novel view synthesis, the recent introduction of NeRF \cite{nerf} model releases the model from the problems of low-resolution geometry and photo-unrealistic rendering of novel views. Instead of explicit representations, NeRF uses an implicit 5D radiance field to represent a scene, which can produce more detailed realistic novel view renderings. Several NeRF-based methods \cite{nerf--, nerfwild} also free the NeRF model from camera parameters. In addition, some methods \cite{dnerf, dmv, hdnerf} also equip NeRF with the capacity to render novel views in dynamic scenes. With implicit density and color prediction of samples on the rays cast from each camera center through the collected photos by volume rendering, NeRF-based models have been the most convincing ones for novel view synthesis. Several works~\cite{insnerf, dm, inplace} try to decompose NeRF field with its implicit representation of scene objects. Although exceptional performance is obtained on some synthetic and real indoor scene datasets, the high cost of 2D ground truth mask annotations makes them less useful in real-world scenarios. 


\vspace{-5pt}
\subsection{Open-Vocabulary 2D Segmentation}

To obtain accurate 2D segmentation masks, traditional methods \cite{maskrcnn, maskformer, mask2former, eva, maskdino} achieve excellent results on annotated closed-set public datasets like COCO \cite{coco} and ADE20K \cite{ade20k}, their performance on Out-Of-Distribution (OOD) objects is much poorer.
Some open vocabulary segmentation models \cite{dino, dinov2, lseg, groupvit, openseg} inspired by openclip \cite{openclip} use multi-modal vision-and-language methods to segment OOD objects. They rely on calculating the pixel-wise similarity between the image embeddings and the language embeddings, obtained by image encoders \cite{vtdense, 1616} and language encoders~\cite{openclip}.
%
Although some works \cite{denseopenclip, cris, simple} make progress in the simple scenarios, the under-expected predictions in complex in-the-wild scenes still block their way to real-world applications

Recently, SAM~\cite{sam}, trained on large-scale datasets, has been shown to provide a good solution. SAM uses a transformer-based image encoder for feature extraction, a prompt encoder for query tokenization, and a mask decoder to output the segmentation results. \cite{sam1, sam2, sam3, sam4} show that a pre-trained SAM model can benefit several downstream tasks in diverse applications such as medical image segmentation and 3D decomposition. ~\cite{samglass, samspace, samcount} point out that the model results are unsatisfactory on other tasks such as segmenting images of glasses and those taken from space.

\subsection{2D Features Field Distillation into NeRF.}

Several works \cite{plift, inplace, ffd, n3f} distill the knowledge of off-the-shelf, supervised, and self-supervised 2D image feature extractors into a 3D feature field. Specifically, FFD \cite{ffd} and N3F \cite{n3f} distill the pixel-level embedding vectors from DINO \cite{dino}and LSeg \cite{lseg} into mplicit neural fields and could localize the corresponding objects matched with the given queries such like point-and-click selections, images patches, and texts.
Besides, although these methods provide reasonable results, they support vocabulary queries from only predefined classes, from ADE20K~cite{ade20k} or COCO~\cite{coco} datasets. Recently, Language Embedded Radiance Fields (LERF) \cite{lerf} have been shown to perform better on a broad range of open-vocabulary queries, including combinations of concepts, colors, long-tail words, and text. However, LERF can provide only an approximate location of an object corresponding to a given vocabulary query; it cannot guarantee accurate segmentation. 
%
The current methods struggle to manage open-vocabulary NeRF decomposition well because they exhibit a trade-off between query processing flexibility and segmentation accuracy. Some methods prioritize accuracy at the expense of processing flexibility, making it difficult to handle open-vocabulary queries, while others can process open-vocabulary queries, but at the cost of generating imprecise segmentation results.

\begin{figure*}[t]
  \centering
  \includegraphics[width=.90\linewidth]{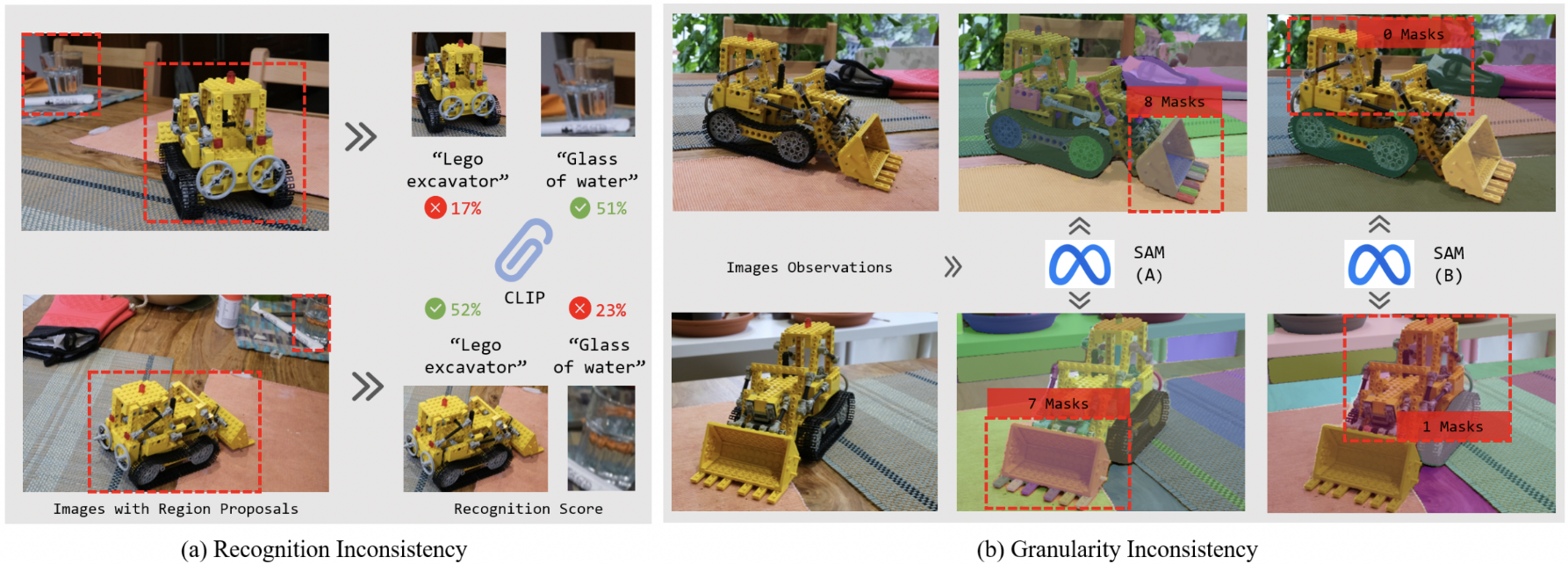}
    \caption{Viewpoint Inconsistency Analysis. (a) Recognition inconsistency of openclip across different viewpoints. The back view of the Lego excavator presents a greater challenge for identification compared to the side view. Additionally, occlusion makes it difficult to identify the glass of water from certain viewpoints. (b) Granularity inconsistency of mask proposals generated by the Segment Anything Model (SAM) across viewpoints. Regions with different colors overlaid on the original image represent the corresponding mask proposals, while regions without colors indicate the absence of mask proposals. Under setting (A), SAM adheres to the default configuration, whereas under setting (B), the hyperparameter \texttt{points\_per\_side} is decreased to $8$. SAM currently lacks the capability to automatically generate a region proposal that accurately encompasses the entire Lego excavator (complex objects) from certain viewpoints. Additionally, SAM may generate different numbers of mask proposals for the same parts in different viewpoints, resulting in granularity inconsistency.}
  \label{fig2}
  \vspace{-5pt}
\end{figure*}

\section{Preliminaries}

\subsection{Neural Radiance Fields}

NeRF~\cite{nerf} utilizes Multilayer Perceptrons (MLPs) to estimate continuous 3D scene geometry and appearance, given a set of images acquired using given camera parameters. It operates with the input of 5D vectors, consisting of point coordinates and viewing directions, denoted by $p = (x, y, z, \theta, \phi)$, and forecasting volume density $\delta$ and color $c = (r, g, b)$ for point, $p$.
%
This scene representation can be visually rendered and optimized using volume rendering techniques. For a pixel's camera ray, defined as $\mathbf{r}(t)=\mathbf{o}+t \mathbf{d}$, with a depth $t$ within bounds $\left[t_{near}, t_{far}\right]$, the camera position $\mathbf{o}$, and viewing direction $\mathbf{d}$, NeRF computes the color of a ray by rendering $N$ sampled points $\left\{\mathbf{x}_n\right\}_{n=1}^N$ with respective depths $\left\{t_n\right\}_{n=1}^N$ as
\begin{equation}
\label{eq1}
\begin{split}
&\hat{\mathbf{C}}(\mathbf{r})=\sum_{n=1}^N \hat{T}\left(t_n\right) \alpha\left(\sigma\left(\mathbf{x}_n\right) \delta_n\right) \mathbf{c}\left(\mathbf{x}_n, \mathbf{d}\right),\\
&\quad \hat{T}\left(t_n\right)=\exp \left(-\sum_{n^{\prime}=1}^{n-1} \sigma\left(\mathbf{x}_{n^{\prime}}\right) \delta_{n^{\prime}}\right),
\end{split}
\vspace{-0pt}
\end{equation}
where $\alpha(x)=1-\exp (-x)$, and $\delta_n=t_{n+1}-t_n$ represents the distance between successive point samples. NeRFs are optimized exclusively on an image dataset and their respective camera parameters by minimizing rendering loss.

\subsection{Segment Anything Model}
As a large ViT-based model trained on an extensive visual corpus (SA-1B), SAM~\cite{sam} exhibits impressive segmentation capabilities across diverse scenarios. There are two primary approaches to harness the capabilities of SAM: (1) Guided by prompts, and (2) Through automatic generation of mask proposals. Both begin by feeding input images into the image encoder. In the first approach, prompts, which may consist of points, a box, a mask, or text, are fed into the prompt encoder. The mask encoder receives the outputs from both the image and the prompt encoders as inputs and generates mask proposals corresponding to these prompts. In contrast, the second approach enables SAM to generate mask proposals without any user prompts. This is accomplished by automatically generating a set of evenly distributed points at a user-specified interval within the images, which are then used as prompts. 


\textbf{Challenges in Utilizing SAM for Open-Vocabulary NeRF Decomposition.} In the realm of open-vocabulary NeRF decomposition, the sole input typically comprises a collection of images alongside corresponding camera parameters. Remarkably, even in the absence of explicit camera parameters, methods like COLMAP~\cite{colmap} can readily extract this information from the input images. With such data at our disposal, the ideal model should not only be capable of synthesizing novel views but also possess the ability to decompose any objects within the scene through open vocabulary prompts.
However, during the training phase, user prompts for SAM are not readily available, necessitating the employment of automatic methods. This strategy, unfortunately, presents its own set of challenges. As demonstrated in Figure \ref{fig2} (b), SAM fails to generate mask proposals as anticipated. Moreover, the granularity of mask proposals across different viewpoints lacks consistency. These issues collectively suggest that directly distilling the knowledge from SAM to 3D fields may not be an optimal solution, underlining the inherent challenges of using SAM for open-vocabulary NeRF decomposition.

\section{Open-vocabulary Embedded Neural Radiance Fields}
In this section, we present our proposed method, Open-vocabulary Embedded Neural Radiance Fields (Open-NeRFs), which addresses the trade-offs in existing methods. Open-NeRFs can flexibly handle open vocabulary queries while generating precise segmentation results for specified objects in NeRF. First, we provide a detailed introduction to our proposed \textbf{integrate-and-distill paradigm}, which ensures recognition and granularity consistency across all viewpoints. Then, we describe the \textbf{hierarchical feature fusion} technique, which enables handling queries of different scales while preventing the loss of local detail. In addition, we explain how to perform \textbf{open-vocabulary querying} using Open-NeRF.

\subsection{Integrate-And-Distill Paradigm}
\label{sec4.1}

As shown in Figure \ref{fig1}, our approach follows the integrate-and-distill paradigm, which contains $3$ main steps: (1) 2D knowledge extraction, (2) 2D knowledge integration across all viewpoints, and (3) integrated knowledge distillation to 3D fields.

\subsubsection{Multi-view Knowledge Integration}
Given a collection of images $\mathbf{I} = \{i_n\}_{n=1}^N$ with corresponding camera parameters, we first leverage the region proposal generator to produce $K_n$, $n\in \{1, ..., N\}$, region proposals $\mathbf{R} = \{r_k\}_{k=1}^{K_n}$, \textit{i.e.}, bounding boxes, for image $i_n$. Then we utilize $\mathbf{R}$ as prompts for SAM to get $K_n$ mask proposals $\mathbf{M} = \{m_k\}_{k=1}^{K_n}$. After that, we crop images along those mask proposals to get $K_n$ sub-images $i^s_{k}$ for $i_n$ as shown in Figure \ref{fig1} and obtain $\sum_{n=1}^N K_n$ sub-images from all $N$ images. Noted that $K_n$ varies between images because some objects are not visible in some viewpoints. Then we align all the sub-images of the same object from different viewpoints with the help of tracking models. For each object $\mathbf{O}_j$, there exists a set of sub-images $\mathbf{I}^s_j = \{i^s_{j, 1}, ..., i^s_{j, L}\}$, where $j$ is the index of objects within the 3D scene and $L$ is the frequency of $\mathbf{O}_j$ being included in the images. We then feed $\mathbf{I}^s_j$ to the image encoder of openclip~\cite{openclip} and a norm operator to get a set of normalized image embeddings $E_j = \{e_{j, 1}, ..., e_{j, L}\}$ for each object and integrate the knowledge from them by fusing $E_j$. 
One simple way of fusing is averaging all image embeddings from the same object: $E^f_j = \frac{1}{L} \sum_{l = 1}^L e_{j, l}$. But this cannot ensure that the norm of $E^f_j$ is $1$, which will introduce errors when calculating similarity with the normalized text embeddings in the querying step. Therefore, we obtain fused embeddings by the following equation: $E^f_j = \sum_{l = 1}^L e_{j, l}/|\sum_{l = 1}^L e_{j, l}|$. With the per-object fused embeddings $E^f_j$ and the mask proposals $m_j$ for each object across images, we obtain per-pixel image embeddings $\mathbf{E} \in \mathbb{R}^{N \times H \times W}$ by assigning all pixels from the same mask proposal $m_j$ the corresponding $E^f_j$, where $(H, W)$ denote (height, width) of $N$ input images.

\begin{figure*}
  \centering
  \includegraphics[width=.95\linewidth]{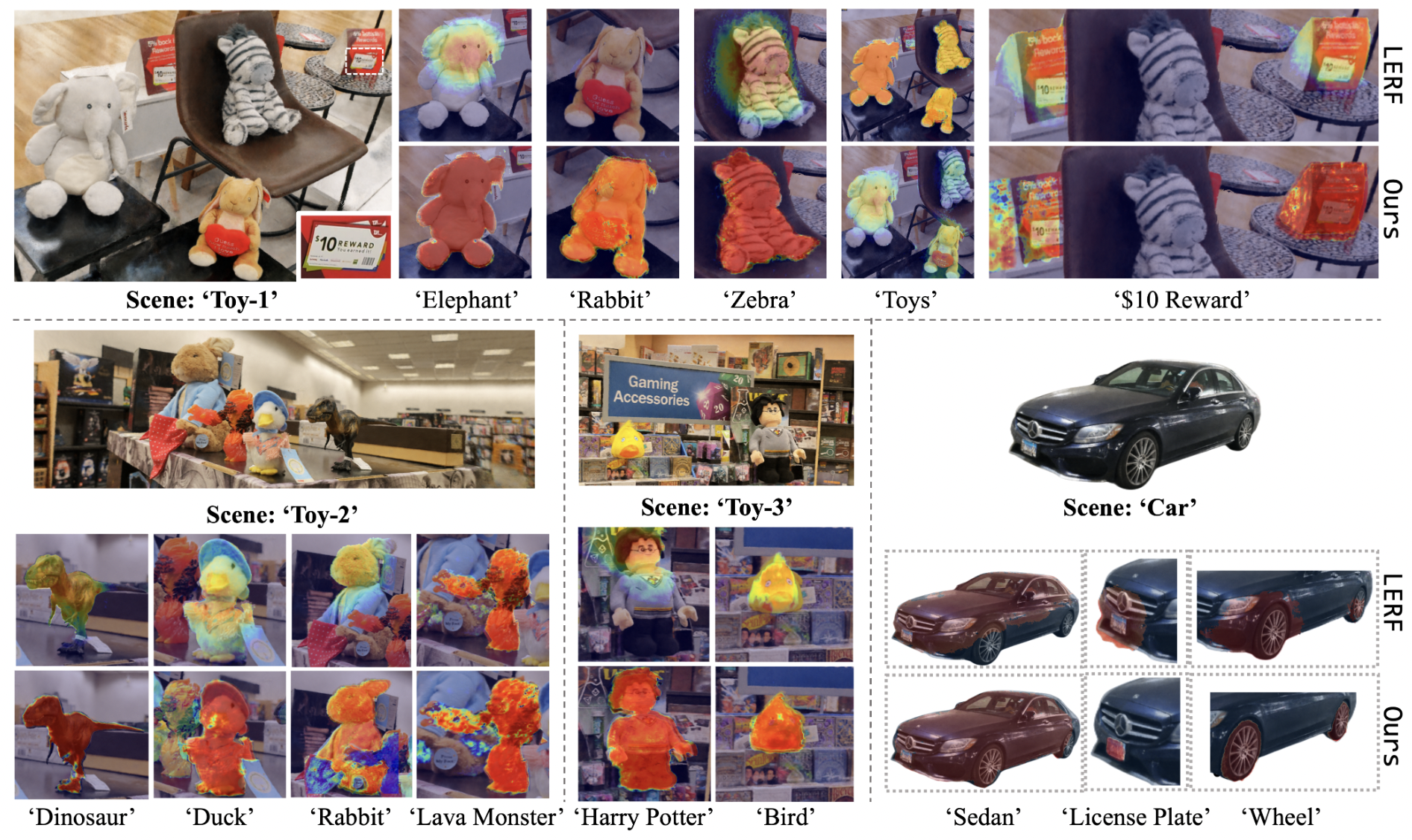}

    \caption{Relevance scores obtained by Open-NeRF and LERF in the scenes: \textit{'Toy-1,2,3'} and \textit{'Car'}.}
  \label{fig5}
\end{figure*}

\subsubsection{Proposals Alignment}

For aligning the mask proposals with different views, we leverage the tracking model to track each object and region of the background and provide a per-pixel prediction of their IDs: $\mathbf{I} \in \mathbb{R}^{H \times W}$ for every frame. Because we demonstrate that the segmentation results from the tracking model are not reliable across frames, we count the ID of each pixel in the mask proposal $\mathbf{M}$ and choose the one with the highest frequency within the entire mask proposal instead of directly using the segmentation results from the tracking model as mask proposals. Then the ID predictions are used to align the same object from different view points.

\subsubsection{Knowledge Integration through Open-Vocabulary Embedding Field}

NeRF-based models employ neural radiance fields to determine view-independent volume density $\delta(x)$ and view-dependent color $c(x, d)$. Some variants of NeRF further expand upon this by incorporating an auxiliary decoder designed to predict additional properties of interest. For instance, Semantic-NeRF~\cite{inplace} utilizes an extra branch to estimate the probability distribution of closed-set semantic labels, DM-NeRF~\cite{dm} learns a unique code for each object within predefined classes, and FFD~\cite{ffd} introduces a feature branch to generate a feature vector corresponding to the 3D coordinates.

Drawing upon these advancements, we propose an additional branch which we call the "open-vocabulary embedding field", designed to facilitate learning of open-vocabulary embeddings. During the multi-view knowledge integration phase, we produce per-pixel image embeddings $\mathbf{E}$ that encapsulate the integrated 2D knowledge aggregated from multiple perspectives. This knowledge is subsequently distilled into the open field. For any given 3D coordinate $\mathbf{x}_n$, where $n\in \{1, ..., N\}$ and $N$ indicates the number of points on the ray $\mathbf{r}$, the open field yields a field embedding $\mathbf{e}(\mathbf{x_n})$, as illustrated in Figure \ref{fig3}. Analogous to volume rendering in NeRF as depicted in Eq.\ref{eq1}, we extract the predicted per-pixel field embeddings through volume rendering along a set of rays. For each individual ray $\mathbf{r}$, the computation is performed as follows:

\begin{equation}
\label{eq2}
\hat{\mathbf{E}}(\mathbf{r})=\sum_{n=1}^N \hat{T}\left(t_n\right) \alpha\left(\sigma\left(\mathbf{x}_n\right) \delta_n\right) \mathbf{e}\left(\mathbf{x}_n\right),
\end{equation}

Note that in contrast to the color function $\mathbf{c}$, the function $\mathbf{e}$, akin to the density function $\sigma$, is direction-independent, thereby ensuring that the semantics of a specific object remains invariant to viewing point changes.

We optimize $\mathbf{e}$ by minimizing the discrepancy between the rendered embedding $\hat{\mathbf{E}}(\mathbf{r})$ and the corresponding per-pixel image embeddings $\mathbf{E}[h,w]$, where $h,w$ denote the pixel's position correlating to the ray $\mathbf{r}$. Consequently, the composite loss integrates both the photometric loss and the distillation loss:
$L=L_p+\lambda L_e,$
where, $L_p$ signifies the Mean Squared Error (MSE) loss between the ground truth color and the rendered color, whereas $L_e$ denotes the mean Huber loss between $\mathbf{E}[h,w]$ and $\hat{\mathbf{E}}(\mathbf{r})$. By default, $\lambda$ is set to $0.1$. To circumvent the potential disruption caused by external embedding branches to the original NeRF, we ensure the embedding field operates independently from the original NeRF, thereby avoiding mutual interference. As a result, $L_e$ is solely employed for optimizing the embedding field.

\subsubsection{Hierarchical Embedding}

We present a novel approach Open-NeRF that capitalizes on hierarchical embeddings, enabling the decomposition of the 3D field in a hierarchical manner based on the scale of input queries. Our method involves the extraction of image embeddings at three distinct levels: (1) object level, (2) part level, and (3) background level.
For the object and part level, we employ region proposal generation techniques at different granularity to identify and propose regions corresponding to complete objects. Following the methodology outlined in Sec \ref{sec4.1}, we extract features for each identified object or object part and embed them into the 3D field. During inference, Open-NeRF provides $2$ relevancy predictions for object level and part level respectively, and automatically selects one following the rule: select the part level predictions only when there are at least $N$ pixels with a higher relevancy score than the max relevancy score of object level, and we set $N=100$ by default.
However, we encounter challenges when processing background regions, such as \textit{grass, road,} and \textit{sky}, as they typically encompass a substantial portion of the entire image. Consequently, conventional region proposal generators struggle to effectively handle such backgrounds. To address this, we adopt an alternative approach for the background component. Instead of employing the region proposal paradigm, we directly utilize a per-pixel feature extractor, such as LSeg~\cite{lseg}. Due to the comprehensive coverage of background categories in the training set of LSeg, it yields decent results for background. However, for open-world objects, it struggles to deliver accurate segmentation outcomes, as demonstrated in the appendix.
In summary, our proposed method involves the extraction of per-pixel embeddings for each object and object part. We assign these embeddings to the pixels within the corresponding mask proposals. For pixels that remain unassigned, we utilize the embeddings generated by openclip-LSeg~\cite{lseg} directly. By combining these hierarchical embeddings, we achieve a hierarchical representation of the 3D field in Open-NeRF. 

\begin{figure*}[htp]
  \centering
  \includegraphics[width=0.8\linewidth]{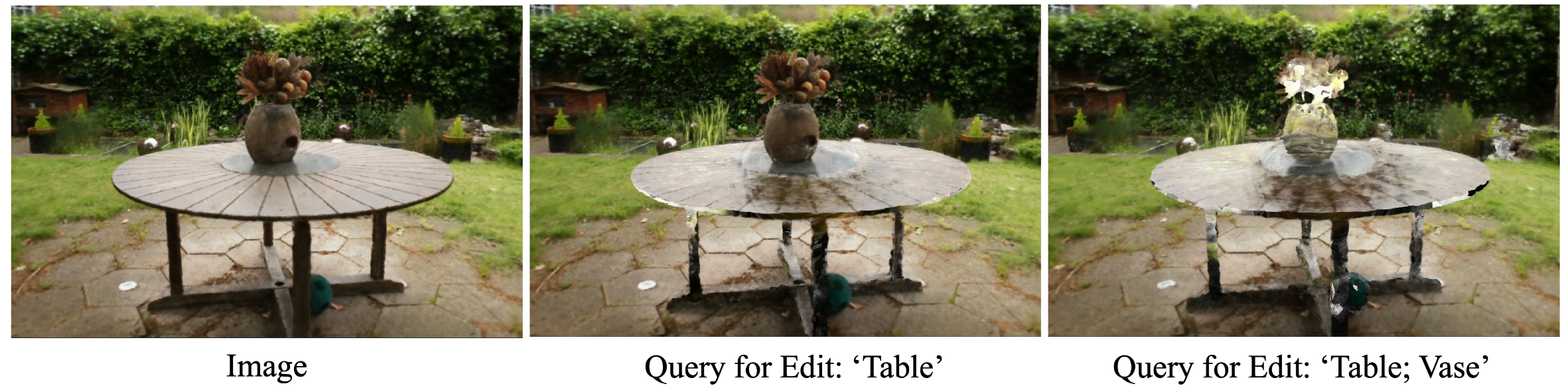}
  \caption{Query-guided texture modifications based on the NeRF decomposition results from our proposed approach: Open-NeRF. The results are achieved by only giving a query for the target object and a query for the modification. For the editing step, we follow the CLIP-NeRF \cite{clipnerf}.}
  \label{figsm3}
\end{figure*}

\subsection{Open-vocabulary Querying}

To enable open-vocabulary querying in NeRF decomposition, a model must be capable of predicting relevancy scores for given coordinates based on the open vocabulary queries received. A threshold can then be employed on the relevancy score to facilitate decomposition. Given a query $\mathbf{q}$, we initially utilize the text encoder of openclip to obtain the normalized text embedding $\mathbf{e}_t(\mathbf{q})$ for $\mathbf{q}$. As mentioned earlier, Open-NeRF generates hierarchical embeddings $\mathbf{e}(x_n)$ for each coordinate $x_n$. Consequently, we calculate the relevancy score $S_r(x_n)$ using the following equation: $S_r(x_n) = \mathbf{e}(x_n) \cdot \mathbf{e}_t(\mathbf{q})$.
For NeRF decomposition, a threshold can be employed to determine whether some points belong to the object based on the given query. To generate 2D segmentation results for novel views, we render the field embeddings along the ray $\mathbf{r}$ following Eq.\ref{eq2}, resulting in the rendered embedding $\mathbf{e_r}(\mathbf{r})$. Then, we calculate the relevancy score for the pixel corresponding to $\mathbf{r}$ through the dot product between $\mathbf{e_r}(\mathbf{r})$ and the text embedding $\mathbf{e}_t(\mathbf{q})$.
As discussed in the preceding section, we learn embeddings at the object level, part level, and background level. Hence, during querying, the model can select the optimal level by comparing the maximum relevancy scores computed using the embeddings from these three levels.


\begin{table*}[]
\centering
\scalebox{0.72}{
\begin{tabular}{cc|ccccc|ccc|cccccc}
\bottomrule[2pt]
\multirow{2}{*}{Method}                                                         & \multirow{2}{*}{Metric} & \multicolumn{5}{c|}{Scene: \textit{Toy}-1}              & \multicolumn{3}{c|}{Scene: \textit{Kitchen}}  & \multicolumn{6}{c}{Scene: \textit{Garden}}\\ \cline{3-16} 
                                                                                &                         & Elephant & Rabbit & Zebra & Reward & Mean & Lego   & Glasses   & Mean  & Vase & Table & Grass & Foot Ball   & Umbrella   & Mean \\ \hline
\multirow{2}{*}{LERF}                                                           
                                                                                & AUPRC                   & 84.7     & 20.1   & 76.3  & 84.9   & 66.5 & 71.3   & 40.5      & 55.9  & 74.1  & 78.3   & 65.6 & 75.1   & 58.3      & 70.3  \\
                                                                                & FPR$_{95}$                     & 5.4      & 40.3   & 1.9   & 0.5    & 12.0 & 10.2   & 33.1      & 21.7  & 8.4  & 7.3   & 18.3 & 8.1   & 43.1      & 17.1 \\\hline
\multirow{2}{*}{Ours}                                                          
                                                                                & AUPRC                   & 96.1     & 88.7   & 97.4  & 89.1   & 92.8 \color{green}$_{\uparrow 26.3}$ & 83.1   & 78.2      & 80.7 \color{green}$_{\uparrow 24.8}$ & 85.8  & 90.4   & 70.2 & 91.3   & 62.8      & 80.1 \color{green}$_{\uparrow 9.8}$ \\
                                                                                & FPR$_{95}$                     & 1.1      & 3.6    & 0.07  & 0.2    & 1.2 \color{green}$_{\uparrow 10.8}$ & 6.1    & 7.4       & 6.8 \color{green}$_{\uparrow 14.9}$ & 4.9  & 3.3   & 8.5 & 2.1   & 21.7      & 8.1 \color{green}$_{\uparrow 9}$  \\\hline
\multirow{2}{*}{\begin{tabular}[c]{@{}c@{}}Ours \small{w/o}\\ \small{Integration}\end{tabular}} & AUPRC                   & 95.1     & 86.3   & 96.9  & 88.2   & 91.6 & 70.1   & 63.1      & 66.6  & 79.3  & 87.1   & 68.4 & 87.7   & 60.9      & 76.7 \\
                                                                                & FPR$_{95}$                     & 1.7      & 4.1    & 0.11  & 0.23   & 1.54 & 10.3   & 18.1      & 14.2  & 6.8  & 5.3   & 9.2 & 3.4   & 20.1      & 9.0 \\ \hline
\end{tabular}}
\scalebox{0.75}{
\begin{tabular}{c|c|ccccc|cccc|cccc}
\hline
\multirow{2}{*}{Method} & \multirow{2}{*}{Metric} & \multicolumn{5}{c|}{Scene: Toy-2}                                                                & \multicolumn{4}{c|}{Scene: Toy-3}                                                                                                               & \multicolumn{4}{c}{Scene: Car}                                                        \\ \cline{3-15} 
                        &                         & Dinosaur & Duck & Rabbit & \begin{tabular}[c]{@{}c@{}}Lava\\ Monster\end{tabular} & Mean         & \begin{tabular}[c]{@{}c@{}}Harry\\ Potter\end{tabular} & Bird & \begin{tabular}[c]{@{}c@{}}'Gaming\\ Accessories'\end{tabular} & Mean         & Sedan & \begin{tabular}[c]{@{}c@{}}License\\ Plate\end{tabular} & Wheel & Mean         \\ \hline
\multirow{2}{*}{LEFR}   & AUPRC                   & 78.2     & 95.8 & 79.2   & 21.3                                                   & 68.6         & 60.1                                                     & 96.5 & 53.8                                                           & 70.1         & 73.5  & 45.7                                                    & 53.2  & 57.5         \\
                        & FPR$_{95}$              & 3.1      & 1.4  & 2.3    & 47.0                                                   & 13.45        & 27.3                                                     & 1.2  & 33.1                                                           & 20.5         & 22.6  & 33.3                                                    & 29.4  & 28.4         \\ \hline
\multirow{2}{*}{Ours}   & AUPRC                   & 91.3     & 97.1 & 88.7   & 86.5                                                   & 90.9 \color{green}$_{\uparrow 22.3}$ & 91.2                                                     & 95.3 & 85.2                                                           & 93.3 \color{green}$_{\uparrow 23.2}$ & 93.5  & 85.2                                                    & 90.3  & 89.7 \color{green}$_{\uparrow 32.2}$ \\
                        & FPR$_{95}$              & 2.2      & 0.7  & 3.4    & 3.8                                                    & 2.5 \color{green}$_{\uparrow 10.9}$  & 1.6                                                      & 1.3  & 5.8                                                            & 1.5 \color{green}$_{\uparrow 19}$    & 1.1   & 2.3                                                     & 1.8   & 1.7 \color{green}$_{\uparrow 26.7}$  \\ \bottomrule[2pt]
\end{tabular}}

\caption{Quantitative Results in Scenes \textit{Toy}-1, 2, 3, \textit{Kitchen}, \textit{Garden}, and \textit{Car}. \textit{Kitchen} and \textit{Garden} scenes are provided by Mip-nerf 360~\cite{mip360}.}
\label{table1}
\end{table*}

\section{Experiment}

In this section, we thoroughly evaluate the capabilities of Open-NeRF and perform a comprehensive comparison with the current state-of-the-art method, LERF~\cite{lerf}, as well as methods based on openclip-LSeg, such as FFD~\cite{ffd}. Our evaluation focuses on showcasing the proficiency of Open-NeRF in processing open-vocabulary queries. To achieve this, we conduct experiments not only on established datasets~\cite{mip360} but also on a collection of 5 diverse in-the-wild scenes encompassing office environments, markets, bookstores, and natural landscapes, featuring numerous long-tail objects. Our collected datasets will be introduced in detail in the appendix. All codes, datasets, and more results including videos will be released soon.



\begin{table*}[t]
\centering
\scalebox{0.75}{
\begin{tabular}{ccccccccccc}
\bottomrule[2pt]
\multirow{2}{*}{Method} & \multirow{2}{*}{Metric} & \multicolumn{9}{c}{Scene: \textit{Desktop}}                                                                                                                                                                                                                                                                                                                                          \\ \cline{3-11} 
                        &                         & MacBook & \begin{tabular}[c]{@{}c@{}}Magic \\ Cube\end{tabular}  & iPhone                                                 & \begin{tabular}[c]{@{}c@{}}VR \\ Glasses\end{tabular} & \begin{tabular}[c]{@{}c@{}}Wireless \\ Mouse\end{tabular} & Speaker & Wallet                                                 & \multicolumn{1}{l}{iPad}                              & Mean  \\ \hline
\multirow{2}{*}{LERF}   & AUPRC                   & 88.5    & 91.3                                                   & 83.3                                                   & 73.6                                                  & 34.1                                                      & 84.6    & 88.3                                                   & 76.1                                                  & --    \\
                        & FPR                     & 3.4     & 1.6                                                    & 4.9                                                    & 11.7                                                  & 42                                                        & 5.7     & 3.1                                                    & 8.4                                                   & --    \\ \hline
\multirow{2}{*}{Ours}   & AUPRC                   & 89.2    & 92.6                                                   & 94.5                                                   & 91.2                                                  & 93.2                                                      & 84.1    & 93.9                                                   & 83.7                                                  & --    \\
                        & FPR                     & 2.7     & 1.3                                                    & 0.3                                                    & 2.1                                                   & 1.3                                                       & 5.3     & 0.5                                                    & 3.7                                                   & --    \\ \hline
                        &                         & UNO     & \begin{tabular}[c]{@{}c@{}}Apple \\ Watch\end{tabular} & \begin{tabular}[c]{@{}c@{}}Airpods\\ Case\end{tabular} & Calculator                                            & Lipstick                                                  & Toy     & \begin{tabular}[c]{@{}c@{}}Airpods \\ Box\end{tabular} & \begin{tabular}[c]{@{}c@{}}Yellow \\ Box\end{tabular} &       \\ \hline
\multirow{2}{*}{LERF}   & AUPRC                   & 83.6    & 67.8                                                   & 90.1                                                   & 81.3                                                  & 73.9                                                      & 76.1    & 84.8                                                   & 82.6                                                  & 78.8  \\
                        & FPR                     & 5.1     & 21.3                                                   & 3.2                                                    & 7.7                                                   & 13.7                                                      & 19.5    & 3.4                                                    & 7.8                                                   & 10.15 \\ \hline
\multirow{2}{*}{Ours}   & AUPRC                   & 87.7    & 85.1                                                   & 90.1                                                   & 89.3                                                  & 80.9                                                      & 80.8    & 95.1                                                   & 90.2                                                  & 88.9 \color{green}$_{\uparrow 10.1}$ \\
                        & FPR                     & 4.1     & 4.3                                                    & 3.2                                                    & 3.8                                                   & 9.2                                                       & 10.7    & 0.05                                                   & 0.7                                                   & 3.32 \color{green}$_{\uparrow 6.8}$
\\\bottomrule[2pt]
\end{tabular}}
\caption{Quantitative comparison of Open-NeRF and LERF in Scenes \textit{Desktop}.}
\label{tablesm1}
\end{table*}

\subsection{Implementation Details}

We present the implementation of Open-NeRF within the Nerfstudio framework~\cite{nerfstudio}, building upon the advancements made by both LERF~\cite{lerf} and the Nerfacto~\cite{nerfactor} method. For efficient decomposition operations in future stages, we adopt the proposed sampling strategy employed by Nerfacto. Our approach leverages the Openclip~\cite{openclip} model, specifically the \texttt{ViT-B-32-quickgelu} version, which has been trained on the extensive LAION-400M dataset.
To generate region proposals, we employ GroundingDINO~\cite{groundingdino} as our region proposal generator, utilizing a robust \texttt{Swin-B} backbone. Prior to its use in our framework, GroundingDINO has been pre-trained on multiple datasets, including COCO~\cite{coco}, O365, GoldG, Cap4M, OpenImage, ODinW-35, and RefCOCO. For mask generation, we rely on the capabilities of the Segment Anything Model (SAM)~\cite{sam}, utilizing the \texttt{sam-vit-h-4b8939} checkpoint.
To ensure consistent alignment of mask proposals belonging to the same object across various viewpoints, we incorporate the state-of-the-art SAM-Track~\cite{samtrack} technique. This allows us to establish accurate correspondences between masks, enhancing the overall quality of the decomposition process in our Open-NeRF implementation.

\subsection{Comparison with Current Methods}

In this section, we exhaustively compare both qualitative and quantitative results of our method with state-of-the-art methods, LERF~\cite{lerf}, on multiple open-world scenes. Methods based on open clip-LSeg barely provide decent results on scenes with novel objects, which is shown in the appendix.

\textbf{Qualitative Results.} Figure \ref{fig5} present a comprehensive assessment of the NeRF decomposition capabilities of our novel algorithms, Open-NeRF and LERF, across a variety of scenes. Each sub-image within these figures showcases the corresponding relevancy score, computed via the dot product of the rendered embeddings, generated through Open-NeRF, and the embeddings of the provided text queries.

Open-NeRF demonstrates superior performance, accurately segmenting both common and novel objects regardless of viewpoint. LERF, on the other hand, delivers only approximate results. A distinguishing feature of Open-NeRF is its capacity to process open vocabulary as shown in the appendix, locating objects based on a variety of attributes, including product name, brand, color, material, and any text inscriptions. For example, a 'Purse', either through direct reference or by using its brand, such as 'Gucci'. Similarly, it accurately identifies a 'Candy Box' when given the descriptor 'Yellow Box', effectively excluding boxes of other colors.

Moreover, Open-NeRF displays versatility in grouping objects. As evidenced in Figure \ref{fig5}, it can identify multiple objects as a group or independently, contingent on the provided query. For instance, it can locate three toys simultaneously when given the query 'Toys', or individually, by providing a more detailed descriptor for each toy. Also, Open-NeRF can locate the whole car or parts of the car, \textit{i.e.}, wheels, license plate, and mirror. This demonstrates Open-NeRF's adeptness at handling diverse scales and complex object queries.

\textbf{Quantitative Results.} To numerically evaluate the capacity of the model on open-vocabulary NeRF decomposition, we employ three distinct metrics on multiple random novel views: (a) Pixel-wise Area Under the Precision-Recall Curve (AUPRC), (b) Pixel-wise False Positive Rate when True Positive Rate equals 95\% (FPR$_{95}$), and (c) Area Under the Receiver Operating Characteristic Curve (AUROC). Table \ref{table1} provides quantitative results of our proposed methods Open-NeRF and LERF~\cite{lerf} in 6 scenes. The results clearly indicate that Open-NeRF consistently surpasses LERF across almost all metrics in every test scenario. The superiority of Open-NeRF is particularly notable in the context of long-tail objects or queries. For instance, we achieve $86.5\%$ on AUPRC for 'Lava Monster', surpassing LERF's performance of $21.3\%$ by a significant margin. Similarly, our achievement of $91.2\%$ on AUPRC for 'Harry Potter' showcases a $31.1\%$ increase compared to LERF.
In the appendix, we show the visualization results of our proposed method compared with LERF~\cite{lerf} in the scene: \textit{Desktop} and here we provide the quantitative results as shown in Table.\ref{tablesm1}. Our proposed approach Open-NeRF surpasses LERF in all three metrics. LERF struggles to provide decent results for objects such as 'Wireless Mouse' and 'Apple Watch', where they obtain $42\%$ and $21.3\%$ in FPR$_{95}$, while Open-NeRF achieves $1.3\%$ and $4.3\%$ in FPR$_{95}$ for the same objects.

\textbf{Ablations.} In addition, we conducted an ablation study on the integration procedure, the results of which are presented in Table.\ref{table1}. Incorporating the integration procedure leads to a marked improvement in performance, especially in scenes featuring occluded objects or objects that are illegible from certain perspectives. Notable examples include the 'Glass of Water', which achieves $78.2\%$ on AUPRC with 2D knowledge integration while achieving $63.1\%$ on AUPRC without it, and the 'Lego Excavator' in the \textit{'Kitchen'} scene, which shows $13\%$ improvement on AUPRC after adding the integration procedure. This further underscores the effectiveness of Open-NeRF and its robustness in handling complex and challenging scenarios.

\section{Limitations \& Conclusion}
In closing, Open-vocabulary Embedded Neural Radiance Fields (Open-NeRFs), provide a solution to the challenges inherent in open-vocabulary NeRF decomposition. With the innovative integrate-and-distill paradigm and hierarchical embedding, Open-NeRFs facilitate consistent recognition and granularity, irrespective of differing viewpoints by leveraging multiple off-the-shelf models. This approach outstrips current state-of-the-art methods in open-vocabulary scenarios, thereby showcasing its potential for practical applications in fields such as robotics and vision-language interaction with 3D scenes. Importantly, as Open-NeRF is constructed upon the foundation of models like Openclip and SAM, its performance is intrinsically linked to its capabilities. Therefore, as these foundational models continue to improve, we can anticipate a corresponding enhancement in the performance of Open-NeRF.

\clearpage
{\small
\bibliographystyle{ieee_fullname}
\bibliography{egbib}
}

\clearpage
\section{Appendix}
In this section, we introduce further details that, due to space constraints, we were unable to incorporate into the main body of the paper. The expanded discussions are categorized into four parts.
\textbf{(1)} We introduce the novel scenes, which are captured from a diverse range of scenarios, along with examples of our annotations. The purpose of these is to quantitatively evaluate the capability of our model in managing the NeRF decomposition with open vocabularies.
\textbf{(2)} In the main manuscript, we provide the quantitative comparisons between our proposed approach, Open-NeRF, and LERF\cite{lerf} within the \textit{Desktop} scene context, and here we deliver more qualitative comparisons. The comparison underlines the superior performance of our method over its counterpart.
\textbf{(3)} As highlighted in the main manuscript, the open-vocabulary NeRF decomposition capacity of Open-NeRF opens up possibilities for practical applications, particularly in the realm of NeRF editing. To substantiate this assertion, we showcase the results of NeRF decomposition achieved through our approach. Furthermore, we present the outcomes of texture modifications resulting from Open-NeRF outputs, thereby providing tangible evidence of the practical usefulness of our method.
\textbf{(4)} We reveal the results generated through the use of LSeg~\cite{lseg} and show the shortcomings that LSeg is challenged with scenarios that extend beyond its conventional realm as mentioned in the main manuscript.


\section{Proposed Novel Scenes}

The evaluation dataset we introduce is compiled from a diverse range of 8K video frames sourced from mobile phones. Every scene within the dataset contains between 150 to 300 image observations, seized from varying viewpoints with distinct scene coverage angle ranges. To further aid the evaluation of decomposition accuracy, we deliver ground truth 2D segmentation for all new scenes collected by us and also the data from Mip-nerf 360 \cite{mip360}. And we provide an annotation every 10 frames. As shown in Figure \ref{figsm2}, we labeled multiple objects in each scene. The integration of these unconstrained in-the-wild scenes allows for a more comprehensive and pragmatic appraisal of Open-NeRF's efficacy in addressing diverse and challenging real-world scenarios.

As demonstrated in Figure 4 of the main manuscript, the scene, denoted as \textit{'Desktop'}, includes a plethora of common items such as 'phones', 'computers', and 'boxes', in addition to less common objects like 'VR glasses', all housed in a moderately dense space. The setup provides a platform to assess the model's proficiency in the accurate decomposition of an open-vocabulary query within a reasonably congested context.

Figure 5, as presented in the main manuscript, illustrates examples of the \textit{'Toy-1,2,3'} and \textit{'Car'} scenes. The \textit{'Toy'} scene categories encompass five divergent scenarios with a variety of toys, including 'Plush Toys', 'Legos', and 'PVC Figuarts', each situated in distinct environments. These provide long-tail samples where each object may correlate with different textual queries, enabling an assessment of the model's capacity to interact with novel objects and process open-vocabulary queries.

The \textit{'Car'} category is made up of three separate collections of image observations, each featuring one or more vehicles within the scenes. One such collection was sourced in a garage setting, while the remaining scenes originate from an outdoor environment. These diverse scenarios further facilitate the examination of Open-NeRF's performance under various conditions.

\section{Experiments}

\begin{figure*}
  \centering
  \includegraphics[width=1\linewidth]{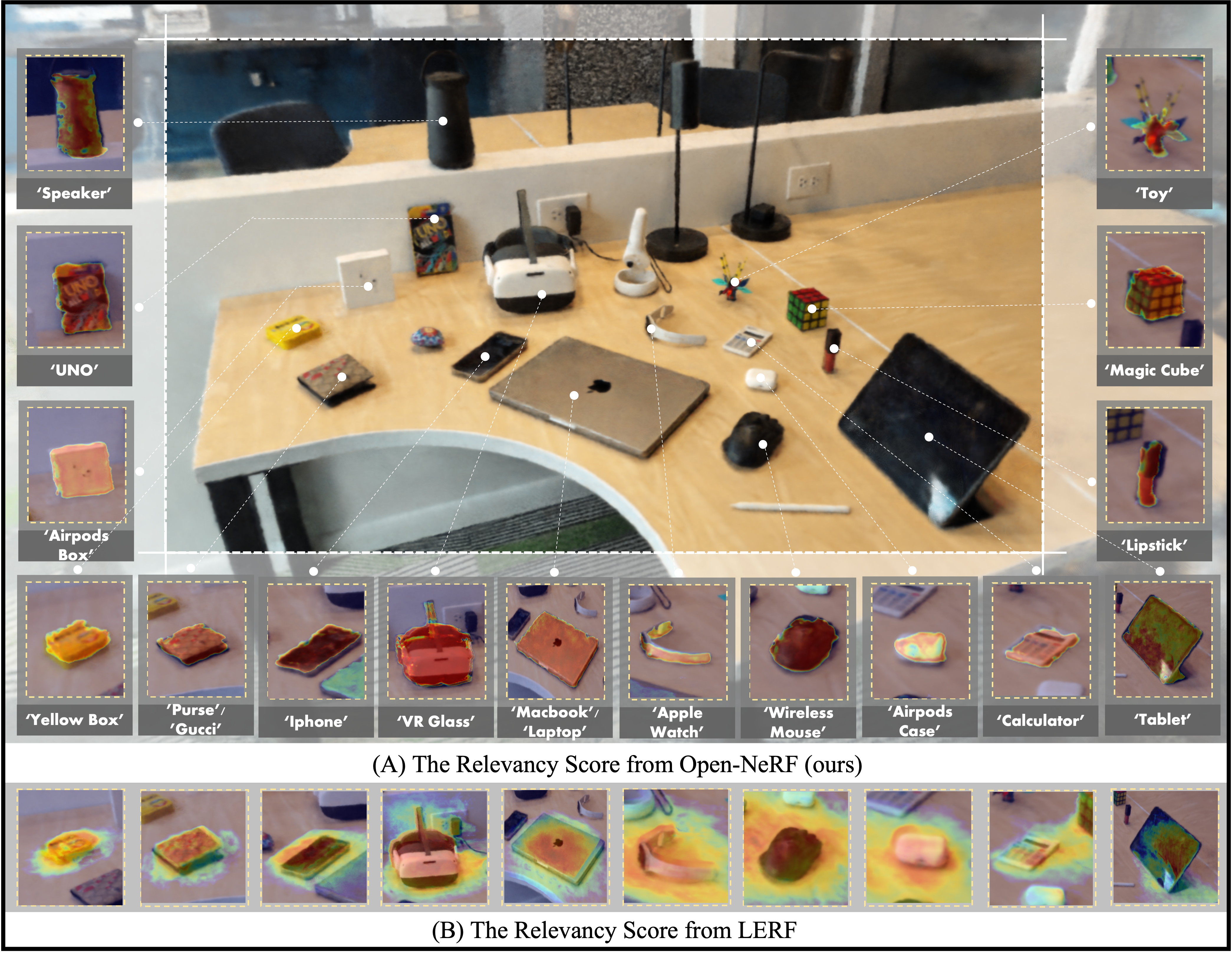}
    \caption{Relevance scores obtained by Open-NeRF and LERF in the \textit{'Desktop'} scene.}
  \label{desktop}
  \vspace{-0pt}
\end{figure*}

\subsection{Quantitative Results}
In the main manuscript, we show the quantitative results of our proposed method compared with LERF~\cite{lerf} in the scene: \textit{Desktop} and here we provide the visual results as shown in Figure  \ref{desktop}. LERF struggles to provide accurate segmentation results for the objects while only providing a rough localization. However, Open-NeRF is able to provide decent 3D segmentation results

\subsection{NeRF Decomposition Results}
\begin{figure}
  \centering
  \includegraphics[width=0.8\linewidth]{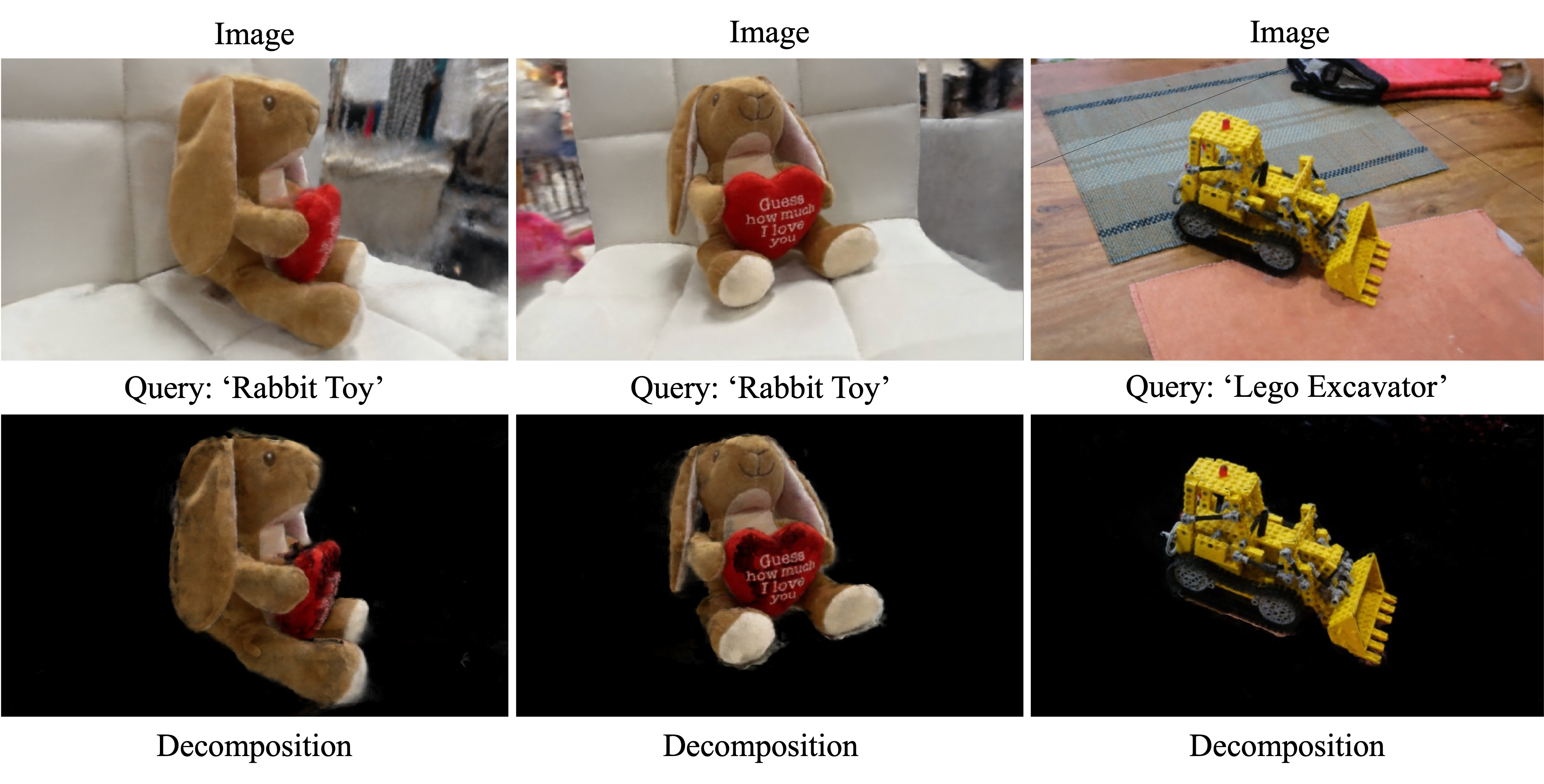}
  \caption{NeRF decomposition results from Open-NeRF with queries: 'rabbit toy' and 'Lego excavator'.}
  \label{figsm4}
\end{figure}

\begin{figure*}
  \centering
  \includegraphics[width=1\linewidth]{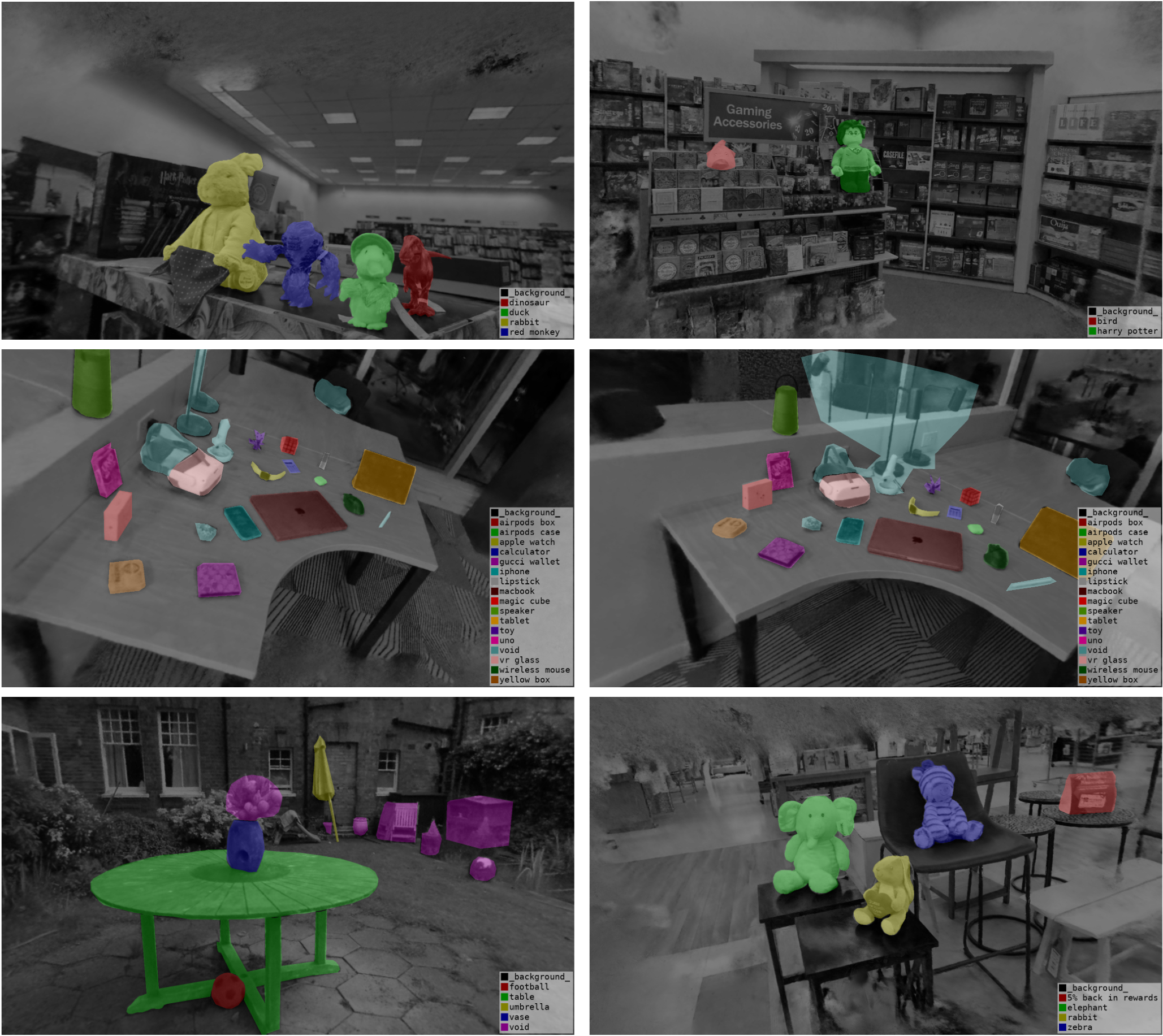}
  \caption{Examples of semantic annotations for the scenes.}
  \label{figsm2}
\end{figure*}

As previously indicated in our main manuscript, Open-NeRF exhibits the capability to generate accurate 3D decomposition results for given open-vocabulary queries. This is visually demonstrated in Figure \ref{figsm4}, which presents the decomposition results for the queries 'Lego excavator' within the \textit{Kitchen} scene, and 'rabbit toy' within the \textit{Toy-4} scene.

Further to this, Figure  5 in the main paper displays the outcomes of texture modifications; specifically, transitioning the textures of 'table' and 'table and vase' to a 'frozen' style\footnote{For the implementation we followed CLIP-NeRF~\cite{clipnerf}.}, based on the results obtained from Open-NeRF. This illustrates the utility of our approach in modifying real-world textures within the 3D decomposition context.

In terms of NeRF decomposition, we employ a threshold strategy for relevancy scores, maintaining only the points that score higher than the predetermined threshold.

\subsection{Qualitative Results of LSeg-based Methods}

In the main manuscript, we present a qualitative and quantitative comparison between our proposed methodology, Open-NeRF, and LERF~\cite{lerf}. The comparative analysis excludes results from methodologies like FFD~\cite{ffd} and N3F~\cite{n3f}. This omission stems from the fact that N3F is predicated on DINO~\cite{dino}, a technique that lacks the capability to align image embeddings with text embeddings originating from open-vocabulary queries. Additionally, the FFD methodology, based on LSeg~\cite{lseg}, is heavily dependent on the performance of 2D CLIP-LSeg. Regrettably, CLIP-LSeg struggles to yield satisfactory outcomes for scenes that contain novel objects or require the processing of open-vocabulary queries.

\begin{figure*}
  \centering
  \includegraphics[width=0.8\linewidth]{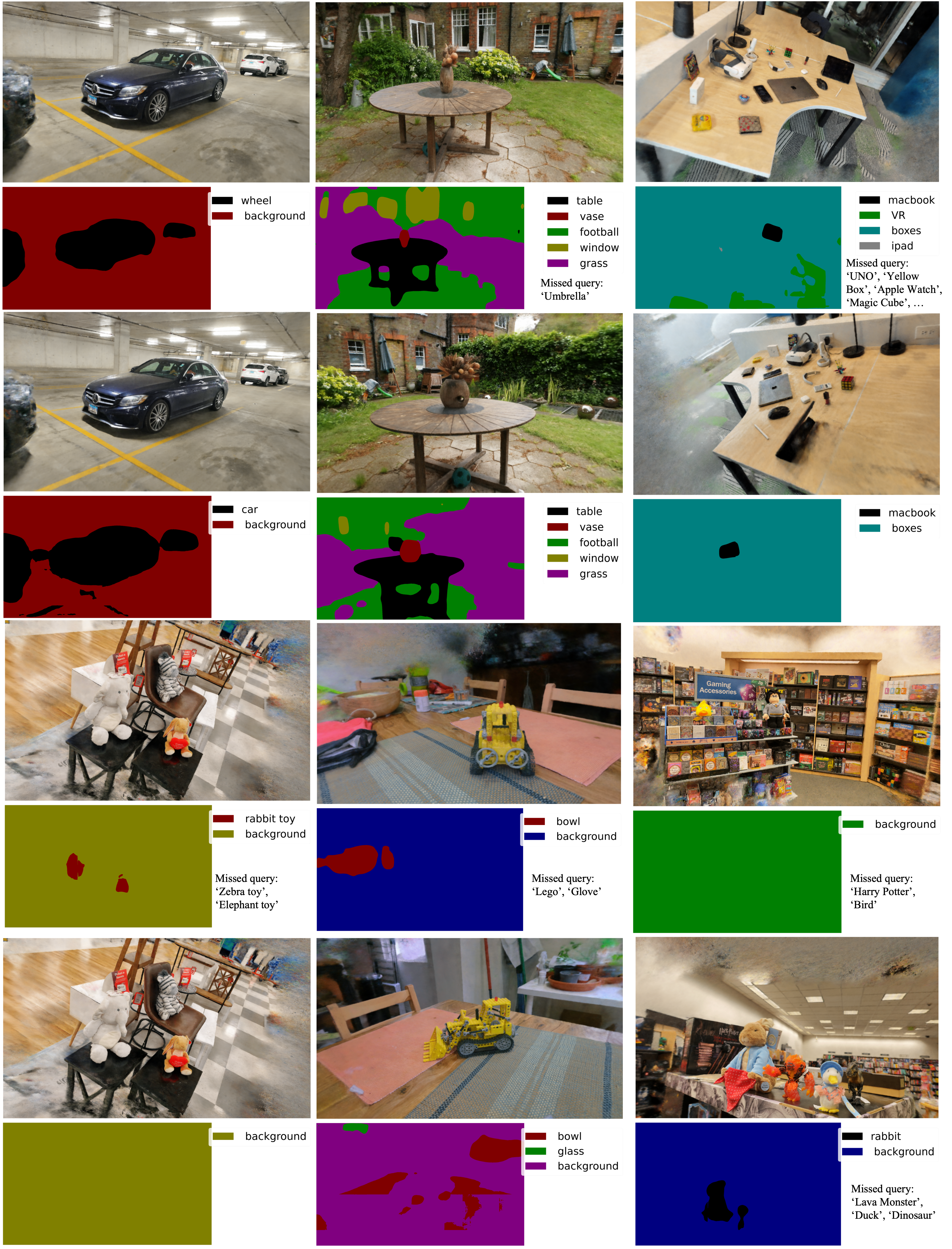}
  \caption{Results of LSeg in 2D open-vocabulary 2D segmentation on multiple scenes.}
  \label{figsm1}
\end{figure*}

As Figure \ref{figsm1} illustrates, CLIP-LSeg can provide acceptable results for common objects within scenes, such as 'table', 'vase', and 'grass' in the \textit{'Garden'} scene sourced from Mip-nerf 360~\cite{mip360}. However, it fails to adequately segment the 'football' within the scene. Moreover, in our collected scenes that comprise a multitude of novel objects such as 'Lava Monster', 'Harry Potter', and 'Lego', CLIP-LSeg consistently fails to yield usable results. It also struggles to handle open-vocabulary queries, providing no reasonable outcomes for queries such as 'Gucci', or 'Yellow Boxes'. Owing to its inability to yield satisfactory results in 2D, the 3D decomposition methodologies premised on LSeg likewise fail to deliver appropriate outcomes for NeRF decomposition.

\end{document}